\documentclass{article}

\usepackage[preprint]{neurips_2025}

\usepackage[utf8]{inputenc} 
\usepackage[T1]{fontenc}    
\usepackage{hyperref}       
\usepackage{url}            
\usepackage{booktabs}       
\usepackage{amsfonts}       
\usepackage{nicefrac}       
\usepackage{microtype}      
\usepackage{enumitem}
\usepackage{tabularx}
\usepackage{geometry}
\usepackage{multirow}
\usepackage[table]{xcolor}
\usepackage{xspace}
\usepackage{amsmath}
\usepackage{siunitx}
\usepackage{graphicx}
\usepackage{cleveref}
\usepackage{arydshln}
\usepackage[most]{tcolorbox}
\usepackage{wrapfig}
\newcounter{boxCounter}

\usepackage{amssymb}

\definecolor{myblue}{HTML}{4E84C4}
\definecolor{myred}{HTML}{B02418}
\definecolor{mygreen}{HTML}{34692E}
\definecolor{myorange}{HTML}{DA7842}
\definecolor{paperblue}{HTML}{077dea}
\definecolor{babyblue}{HTML}{E3EDF7} 

\newcommand\ourmodel{AdaCtrl\xspace}
\newcommand{\coloredalpha}{\textcolor{paperblue}{\alpha}}
\newcommand{\coloredbeta}{\textcolor{paperblue}{\beta}}

\newcommand{\coloredsigma}{\textcolor{paperblue}{\sigma}}

\title{AdaCtrl: Towards Adaptive and Controllable Reasoning via Difficulty-Aware Budgeting}

\author{
  Shijue Huang$^{\coloredalpha}$\thanks{Equal contribution. $^\dag$ Corresponding author.}~,
  Hongru Wang$^{\coloredbeta}$\footnotemark[1]~, 
  Wanjun Zhong\footnotemark[1]~,
  Zhaochen Su$^{\coloredalpha}$,
  Jiazhan Feng$^{\coloredsigma}$, \\
  \textbf{Bowen Cao}$^{\coloredbeta}$,
  \textbf{Yi R. (May) Fung}$^{\coloredalpha}$$^\dag$
  \\
  $^{\coloredalpha}$Hong Kong University of Science and Technology \\$^{\coloredbeta}$ The Chinese University of Hong Kong, 
  $^{\coloredsigma}$ Peking University
}

\begin{document}

\maketitle

\begin{abstract}
Modern large reasoning models demonstrate impressive problem-solving capabilities by employing sophisticated reasoning strategies. However, they often struggle to balance efficiency and effectiveness, frequently generating unnecessarily lengthy reasoning chains for simple problems.
In this work, we propose \ourmodel, a novel framework to support both difficulty-aware adaptive reasoning budget allocation and explicit user control over reasoning depth. \ourmodel dynamically adjusts its reasoning length based on self-assessed problem difficulty, while also allowing users to manually control the budget to prioritize either efficiency or effectiveness.
This is achieved through a two-stage training pipeline: an initial cold-start fine-tuning phase to instill the ability to self-aware difficulty and adjust reasoning budget, followed by a difficulty-aware reinforcement learning (RL) stage that refines the model’s adaptive reasoning strategies and calibrates its difficulty assessments based on its evolving capabilities during online training. To enable intuitive user interaction, we design explicit length-triggered tags that function as a natural interface for budget control.
Empirical results show that
\ourmodel adapts reasoning length based on estimated difficulty,
compared to the standard training baseline that also incorporates fine-tuning and RL, 
it yields performance improvements and simultaneously reduces response length by 10.06\% and 12.14\% on the more challenging AIME2024 and AIME2025 datasets, which require elaborate reasoning, and by 62.05\% and 91.04\% on the MATH500 and GSM8K datasets, where more concise responses are sufficient. Furthermore, \ourmodel enables precise user control over the reasoning budget, allowing for tailored responses to meet specific needs.
Further analysis also reveals that \ourmodel accurately estimates problem difficulty and allocates reasoning budgets in alignment with these assessments\footnote{This work is ongoing, and the code will be released at \url{https://github.com/JoeYing1019/AdaCtrl}.}.
\end{abstract}

\section{Introduction}

With the emergence of test-time scaling~\citep{snell2024scalingllmtesttimecompute,muennighoff2025s1simpletesttimescaling,balachandran2025inferencetimescalingcomplextasks,zhang2025surveytesttimescalinglarge}, large reasoning models such as Deepseek R1~\citep{deepseekai2025deepseekr1incentivizingreasoningcapability} and OpenAI o1~\citep{openai2024o1} have demonstrated superior performance across a variety of tasks by thoroughly exploring various reasoning paths prior to generating final answers. While such deep and long reasoning processes can significantly enhance a model's ability to solve complex problems, they inevitably introduce significant inference overhead and lead to unnecessary overthinking on simpler problems~\citep{chen2025think23overthinkingo1like, sui2025stopoverthinkingsurveyefficient}. For instance, even when presented with easy and straightforward questions like ``Evaluate $log_2(64)$'', these models still tend to engage in lengthy chain-of-thought~\citep{wei2022chain}, unnecessarily employing advanced meta-reasoning skills such as planning, reflection, and verification~\citep{Ryan2016Systems1A,li202512surveyreasoning,wang-etal-2025-self-reasoning}. Such reasoning behavior, while beneficial for complex queries, incurs excessive latency and computational costs, negatively affecting user experience~\citep{wang2025harnessing}.

\begin{figure*}[t]
    \centering
\includegraphics[
width=\linewidth]{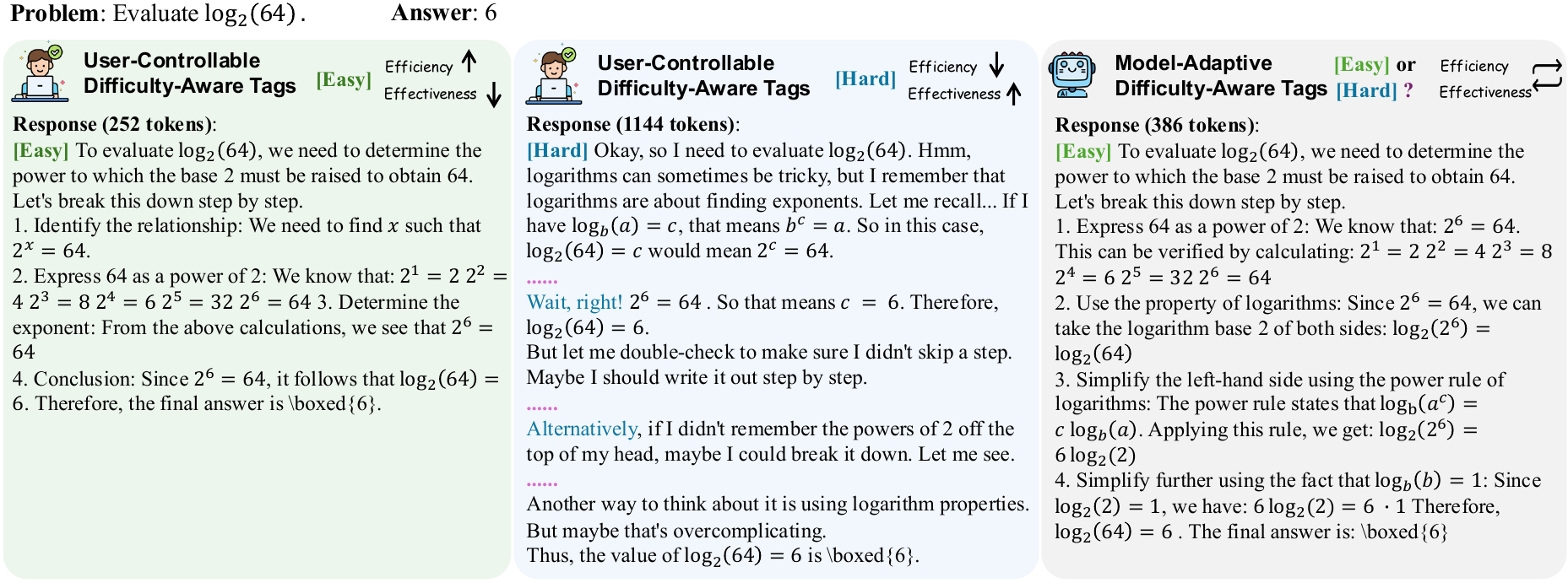}
    \caption{
    Given the same problem, \ourmodel supports three reasoning modes: the easy mode offers concise answers with less tokens; the hard mode delivers extensive responses with more tokens; and the adaptive mode dynamically allocates reasoning budgets according to the problem complexity.
    }
    \label{fig:case}
\end{figure*}

Recent efforts have explored several ways to improve the reasoning efficiency and mitigate overthinking issue. Some approaches aim to minimize reasoning length across all questions, regardless of their actual complexity (e.g., easy or hard), by enforcing conciseness through fine-tuning~\citep{munkhbat2025selftrainingelicitsconcisereasoning, ma2025cotvalvelengthcompressiblechainofthoughttuning} or reinforcement learning~\citep{arora2025traininglanguagemodelsreason, aggarwal2025l1controllinglongreasoning}. However, this universal compression strategy may sacrifice reasoning quality on truly complex tasks that require deeper analysis. On the other hand, some methods attempt to precisely control the token budget for each question, seeking a more efficient yet accurate reasoning process~\citep{nayab2025concisethoughtsimpactoutput, muennighoff2025s1simpletesttimescaling, xu2025chaindraftthinkingfaster}. Nevertheless, such fine-grained control often relies on accurate difficulty assessments and exceptional capabilities of underlying LLMs (e.g., precise instruction-following), which can be inflexible and brittle in practice. It is inherently challenging to determine the optimal token budget beforehand, and models may struggle to reliably interpret or execute fine-grained instructions. These limitations underscore the need for a unified framework that is both \textit{adaptive} and easily \textit{controllable} based on task complexity.

To this end, we propose \ourmodel, a novel framework to 1) allow the model to dynamically adjust their reasoning effort based on self-assessed difficulty (i.e., adaptive reasoning mode); and 2) enable users to specify reasoning mode according to difficulty of problem (i.e., controllable budget allocation). As shown in \figurename~\ref{fig:case}, given a query, \ourmodel offers three reasoning modes. Two of these modes are manually specified to allow users to control the reasoning budget:
(i) Easy, which prioritizes efficiency and provides concise responses to any question;
(ii) Hard, which aims for higher effectiveness by elaborating the full reasoning process and delivering detailed information.
In addition, an adaptive mode automatically adjusts the reasoning effort based on the complexity of the input query, achieving a balance between effectiveness and efficiency without manual intervention.
To achieve this, we start by curating a dataset that covers both \textit{easy} and \textit{hard} subsets and  insert corresponding indicator tags (i.e., ``[Easy]'' or ``[Hard]'') prior to the model’s responses.
Then we utilize cold-start fine-tuning to empower the model with foundation capabilities to estimate the complexity of a question and allocate reasoning budgets accordingly, rendering self-aware difficulty estimation. 
Moreover, we adapt difficulty-aware reinforcement learning framework with carefully designed rewards to calibrate its self-assessment of problem difficulty (i.e., difficulty estimation calibration reward) and to refine the model’s adaptive reasoning behavior (i.e., difficulty-aware length reward).

Experimental results on four benchmarks demonstrate that \ourmodel significantly improves the trade-off between effectiveness and efficiency. It outperforms most baselines across four datasets while efficiently managing the reasoning budget. 
Specifically, compared with the standard SFT + RL baselines,  \ourmodel achieves accuracy improvements of up to 10.14\%, while reducing response length by as much as 91.04\%.
Further analysis demonstrates that \ourmodel offers effective human-in-the-loop controllability via explicit difficulty-aware tags, enables accurate difficulty estimation during reinforcement learning, and maintains robust performance under hyperparameter variation.
Overall, the contributions are as follows:

\begin{itemize}[leftmargin=*]
\setlength{\itemsep}{0pt}
\item We introduce \ourmodel, a unified framework for adaptive and controllable reasoning that supports dynamic trade-offs between efficiency and performance, allowing the model to estimate the difficulty of problem and adjust the reasoning mode itself, and also the user to specify the desired reasoning mode to meet diverse needs in practice.

\item We present a two-stage training paradigm that integrates cold-start fine-tuning and difficulty-aware reinforcement learning together to foster self-awareness of problem difficulty and supports difficulty-aware budget allocation, considering the differences and dynamics
of model capabilities.

\item Empirical results on four benchmark datasets demonstrate that \ourmodel successfully enhance adaptivity and controllability via explicit difficulty-aware tags. Further analysis reveals that \ourmodel serves as an effective difficulty estimator, and accurately enables difficulty-aware budget allocation.

\end{itemize}

\section{Related Work}

\noindent\textbf{Reasoning Efficiency via Supervised Fine-Tuning.}
While LLMs achieve impressive performance on complex tasks by generating elaborate multi-step reasoning chains~\citep{dubey2024llama, su2025openthinkimg}, this capability can lead to excessive verbosity and computational overhead for simpler queries. This ``overthinking'' phenomenon has motivated research into improving reasoning efficiency~\citep{qu2025survey, wang2025harnessing}.
One prominent strategy involves Supervised Fine-Tuning (SFT) to guide models towards more concise reasoning. Some SFT works focus on training with inherently shorter reasoning paths. For example models learn adherence to token budgets through specific prompting during data generation~\citep{han2024token}. Others distill concise paths from best-of-N sampling~\citep{munkhbat2025selftrainingelicitsconcisereasoning} or fine-tune models to omit intermediate steps for samples where the model is already confident~\citep{yu2024distilling21}.
Another SFT direction compresses existing reasoning chains. \citet{kang2024c3ot} employ GPT-4~\citep{achiam2023gpt} as a compressor then fine-tune a model on these long-to-short CoT mappings. LMskip~\citep{liu2024can} induces step-skipping behavior under step constraints. SPIRIT-FT~\citep{cui2025stepwise} identifies critical reasoning steps using perplexity as a guide for pruning. TokenSkip~\citep{xia2025tokenskip} analyzes token importance within CoT outputs for controllable compression. These SFT methods reduce length but often enforce a general conciseness ill-suited for complex problems and typically lack self-assessment of difficulty or user budget control.

\noindent\textbf{Reasoning Efficiency via Reinforcement Learning.}
RL offers another significant avenue for optimizing reasoning efficiency, building upon its success in developing deep reasoning capabilities in models like DeepSeek-Coder~\citep{guo2025deepseek}. Many RL approaches incorporate explicit length-based rewards to encourage conciseness alongside accuracy. Some methods link generation length to task difficulty or directives within the prompt: DAST~\citep{shen2025dastdifficultyadaptiveslowthinkinglarge} adapts CoT length to problem complexity via reward shaping, while LCPO~\citep{aggarwal2025l1controllinglongreasoning} controls length using prompt-specified targets. Other techniques normalize length rewards against baselines, as seen in O1-Pruner~\citep{luo2025o1}, the per-prompt normalization by \citet{arora2025traininglanguagemodelsreason}, and the Kimi 1.5 report~\citep{kimiteam2025kimik15scalingreinforcement}. \citet{yeo2025demystifyinglongchainofthoughtreasoning} proposed a cosine reward to manage length effectively, also highlighting the ``length hacking'' problem where models artificially extend reasoning. Beyond explicit length rewards, alternative RL strategies include meta-RL for test-time optimization~\citep{qu2025optimizingtesttimecomputemeta}, utility maximization for budget awareness~\citep{yu2025think}, preference optimization with heuristics~\citep{chen2025think23overthinkingo1like}, and mitigating GRPO's bias towards longer trajectories~\citep{liu2025understanding}. However, these methods generally lack the explicit user control over reasoning depth offered by \ourmodel's difficulty-aware  tags and do not prioritize training for self-awareness of problem difficulty. Our two-stage SFT-RL framework uniquely addresses these aspects, enabling both autonomous and user-influenced reasoning budgets.

\begin{figure*}[t]
    \centering
\includegraphics[
width=0.9\textwidth]{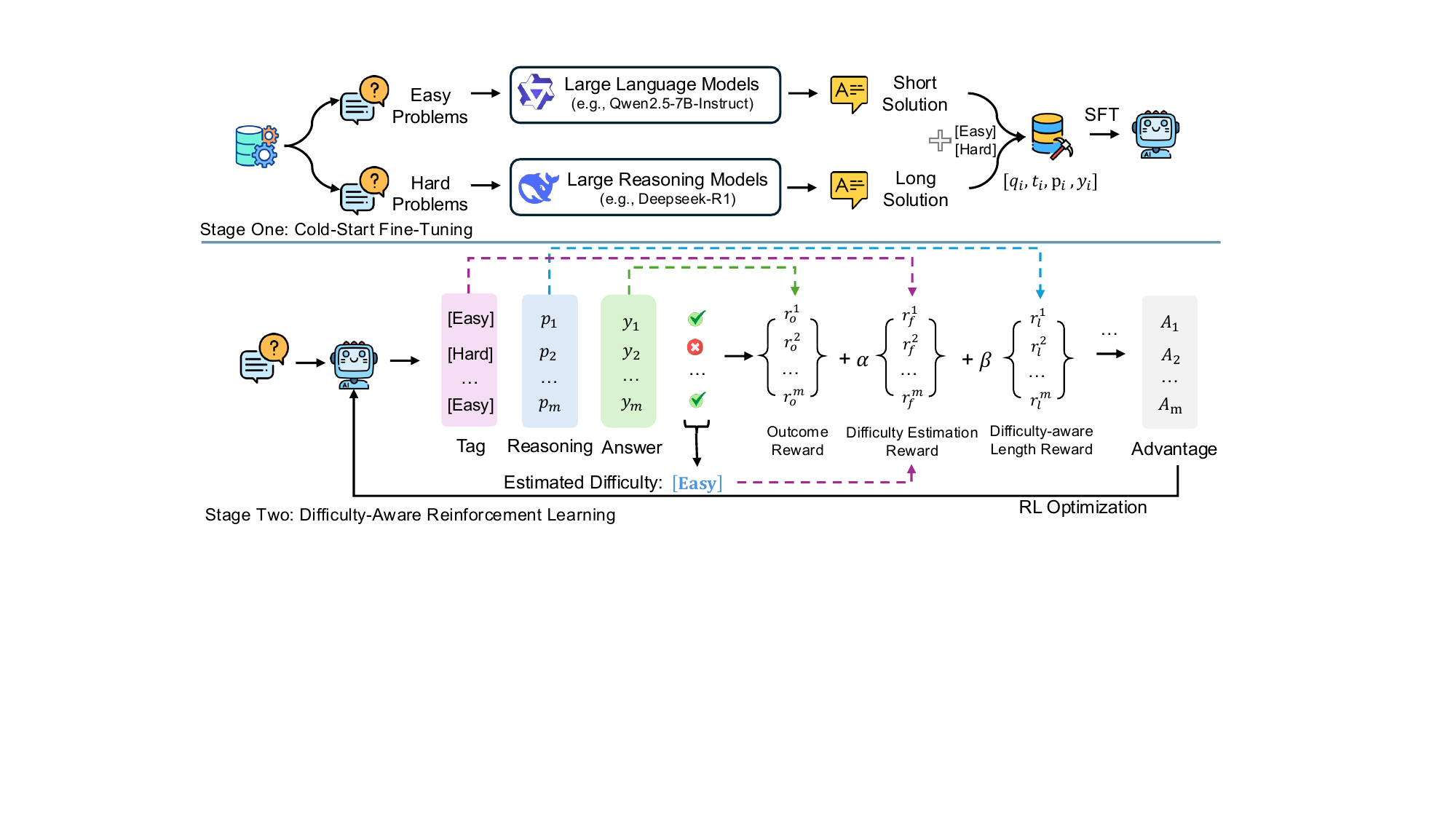}
    \caption{\ourmodel comprises a two-stage training pipeline: the cold-start finetuning first utilizes both short and long reasoning trajectories to establish basic budget awareness; then a difficulty-aware  reinforcement learning framework is utilized to calibrate problem difficulty estimation and develop adaptive reasoning strategies.}
    \label{fig:main}
\end{figure*}

\section{Method}
In this section, we demonstrate the design of our proposed \ourmodel, which includes: (1) Cold-start fine-tuning that provides initialization for difficulty estimation and difficulty-aware budget adjustment; (2) Difficulty-aware reinforcement learning framework that boosts model's capabilities on response length control and difficulty estimation. The entire framework is demonstrated in \figurename~\ref{fig:main}.

\subsection{Cold-Start Fine-Tuning}
This stage primarily focuses on equipping models with the ability to adhere to output formats that include difficulty-aware tags (e.g., "[Easy]'' and "[Hard]'')\footnote{We do not consider more fine-grained category in order to maintain ease of control and ensure reliability.} and to control response length accordingly. To curate suitable training data for this purpose, we directly select both easy and hard problems from the DeepMATH dataset~\citep{he2025deepmath103klargescalechallengingdecontaminated}, which provides difficulty annotations for each problem.
For easy problems $\{q_1^e, q_2^e, ..., q_n^e,\}$, we utilize the instruction model, $\mathcal{M}$,
to
generate concise response, while for hard ones $\{q_1^h, q_2^h, ..., q_m^h,\}$, a strong large reasoning model $\mathcal{R}$ is employed to generate reasoning trace. Then we filter the solutions based on answer correctness for both subsets:
\begin{align}
\mathcal{D}^e = \{(q_i^e, [p_i, y_i]) | [p_i, y_i] = \mathcal{M}(q_i^e), \mathbb{I}(y_i, \hat{y}_i) = 1 \}, \\
\mathcal{D}^h = \{(q_i^h, [p_i, y_i]) | [p_i, y_i] = \mathcal{R}(q_i^h), \mathbb{I}(y_i, \hat{y}_i) = 1 \},
\end{align}
where $ [r_i, y_i]$ is the model-generated response that includes reasoning process $p_i$ and the predicted answer $y_i$,
$\hat y_i$ is the ground truth answer of corresponding samples, and $\mathbb{I}(y_i, \hat y_i) = 1$ represents that the model response $[p_i, y_i]$ is correct with the verification of answer $\hat y_i$. 
Then by prepending the ``[Easy]'' tag 
for response in $\mathcal{D}^e$ and ``[Hard]'' tag for those in $\mathcal{D}^h$, we construct a dataset $\mathcal{D} = \{(q_i, [t_i, p_i, y_i])\}$ for cold-start training, where $t_i$ is the difficulty-aware tag.
After the cold-start finetuning, the model learns to adhere to the specified format and to generate solutions based on its estimated difficulty of each problem, such as allocating more reasoning tokens for hard problems to enable deeper and more diverse reasoning paths. Notably, this also provides greater controllability for non-expert users, who can easily guide the model’s behavior by using simple  tags like “[Easy]” and “[Hard]”.

\subsection{Difficulty-Aware Reinforcement Learning}
After cold-start fine-tuning, we adapt Group Relative Policy Optimization (GRPO)~\citep{shao2024deepseekmathpushinglimitsmathematical} as the reinforcement learning (RL) algorithm because it
produces multiple rollouts that can be naturally utilized to estimate difficulty of given question from the perspective of the trained model. Specifically, we carefully design three types of rewards to guide the difficulty-aware reinforcement learning optimization:

\paragraph{Outcome Accuracy Reward.}

The rule-based outcome accuracy reward 
has been widely utilize in RL training~\citep{guo2025deepseek, aggarwal2025l1controllinglongreasoning, arora2025traininglanguagemodelsreason,yu2025dapoopensourcellmreinforcement,wang2025actingreasoningmoreteaching}, which evaluates the correctness of a generated response:
\begin{align}
    r_o(y_i)=\begin{cases}
1.0& \mathbb{I}(y_i, \hat y_i) = 1\\
-1.0& \text{otherwise}
\end{cases}
\end{align}
To ensure more reliable verification, we explicitly
require the model to present its final answers in a specified format (i.e., within \texttt{\textbackslash boxed\{\}}).

\paragraph{Difficulty Estimation Calibration Reward.}
Accurately estimating problem difficulty is a fundamental aspect of difficulty-aware budgeting. However, since the difficulty of a problem can vary significantly across different models, and obtaining reliable, model-specific difficulty labels is also challenging. To address this, we design a reward function that leverages GRPO's multiple rollouts to estimate a golden difficulty during online training in a natural and effective manner.
Specifically, we
calibrate the estimated difficulty of the model using the frequency of rollouts that leads to a correct answer.
We label a question as easy if the frequency of accurate rollouts exceeds a pre-defined threshold $\delta$, or it is labeled as hard. 
Moreover, during the rollout process, the model is expected to generate a correct difficulty-aware tag at the beginning of the entire response to match the its capability, so the reward function is designed based on the matches between the generated difficulty-aware tag $t_i$ and estimated tag label $\hat t_i$ that is determined by multiple rollouts:
\begin{equation}
    r_f(y_i)=\begin{cases}
1.0& \mathbb{I}(t_i, \hat t_i) = 1\\
0.0& \mathbb{I}(t_i, \hat t_i) = 0\\
-1.0& t_i \text{ cannot be found in } y_i
\end{cases}
\end{equation}

\paragraph{Difficulty-aware Length Reward.}

Different from previous related works that encourage the model to generate concise responses for all problems~\citep{arora2025traininglanguagemodelsreason, aggarwal2025l1controllinglongreasoning}, we hope to encourage such behavior only for easy problems and maintain long thinking capabilities for better tackling hard problems. This helps prevent over-optimization, where the model may fail to engage in deeper thinking when necessary.
To prevent unnecessary overthinking, we design a difficulty-aware length reward that encourage concise responses when the generated difficulty-aware tag $t_i$ is ``[Easy]'':
\begin{equation}
    r_l(y_i)=\begin{cases}
1.0 - \frac{1 - \cos(({l_i^j}/{L_i}) \pi)}{2}& t_i = \text{[Easy]} \\
0.0& \text{otherwise}
\end{cases}
\end{equation}
where $l_i^j$ is the $j$-th rollout length for problem $q_i$, and $L_i$ is the max length in the rollout group of problem $q_i$.
We leverage the monotonicity of the cosine function within a specific domain (i.e., 0–$\pi$). This function is easy to tune and provides smooth behavior, as evidenced by a recent study~\citep{yeo2025demystifyinglongchainofthoughtreasoning}. Therefore, the difficulty-aware length reward assigns a lower score as the generated response length increases.

\paragraph{Overall Reward and Objective.}

To enhance the model's adaptive reasoning capabilities and calibrate its self-evaluation of problem difficulty during online training, the overall reward in the reinforcement learning process is computed by integrating the three specifically designed rewards:
\begin{equation}
    r(y_i) = r_o(y_i) + \alpha \cdot r_f(y_i) + \beta \cdot r_l(y_i)
\end{equation}
where $\alpha$ and $\beta$ are hyper-parameters. During optimization, we sample a collection of problems that covers both easy and hard problems from DeepMATH~\citep{he2025deepmath103klargescalechallengingdecontaminated} to allow the model to learn dynamic strategies for different types of problems. 
Following \citet{yu2025dapoopensourcellmreinforcement,he2025deepmath103klargescalechallengingdecontaminated}, the policy model $\pi_\theta$ is optimized through the following objective:
\begin{small}
\begin{equation}
\begin{aligned}
&\mathcal{J}_\text{GRPO}(\theta) = \mathbb{E}_{(q,a)\sim \mathcal{D}, \{o_i\}_{i=1}^G\sim \pi_{\theta_\text{old}}(\cdot\mid q)} 
\Bigg[ \frac{1}{G}\sum_{i=1}^{G} \frac{1}{|o_i|}\sum_{t=1}^{|o_i|} \Bigg( 
\min \Big( \frac{\pi_{\theta}(o_{i,t} \mid q, o_{i,<t})}{\pi_{\theta_{\text{old}}}(o_{i,t} \mid q,o_{i,<t})}(\theta) \hat{A}_{i,t}, \\& 
\ \text{clip} \Big( \frac{\pi_{\theta}(o_{i,t} \mid q, o_{i,<t})}{\pi_{\theta_{\text{old}}}(o_{i,t} \mid q,o_{i,<t})}(\theta),  1 - \varepsilon, 1 + \varepsilon \Big) \hat{A}_{i,t} \Big)
- \beta' D_{\text{KL}}(\pi_{\theta} || \pi_{\text{ref}}) 
\Bigg) \Bigg],
\label{eq:grpoloss}
\end{aligned}
\end{equation}
\end{small}
where $G$ is the group size, $o_i$ is the rollout response, $\hat{A}_{i,t} $ is the advantage
of the $i$-th response calculated by normalizing rewards in the group, and $\beta', \varepsilon$ are hyper-parameters. Through this reinforcement learning process, models can more effectively assess problem difficulty relative to their own capabilities and develop adaptive reasoning strategies accordingly.

\section{Experiment}

\subsection{Experimental Setup}
\definecolor{mygray}{gray}{0.92}

\textbf{Model and Datasets.}
We adopt Qwen2.5-7B-Instruct and Qwen2.5-14B-Instruct~\citep{qwen2025qwen25technicalreport} as the backbone model for training. During the cold-start fine-tuning phase, the training problems are drawn from the DeepMATH dataset~\citep{he2025deepmath103klargescalechallengingdecontaminated}, which assigns each problem a difficulty level ranging from 1 to 9. 
We categorize problems with difficulty levels of 5 or below as easy, and those above 5 as hard. For easy problems, the backbone model is employed to generate concise responses. In contrast, for hard problems, we incorporate extensive long-form reasoning trajectories generated by Deepseek R1~\citep{deepseekai2025deepseekr1incentivizingreasoningcapability}. After filtering these trajectories using ground-truth answers, we construct a cold-start SFT dataset comprising 8K instances, which includes 4K with short and 4K with long reasoning chains. To support difficulty-aware reinforcement learning, we further sample an additional 30K examples from DeepMATH that comprise 10K easy and 20K hard problems and are distinct from those in the cold-start fine-tuning dataset.

\paragraph{Baselines.} To assess the effectiveness of \ourmodel, we compare it against several baselines that share the same backbone model. These include:
(1) Base Model: the unmodified base instruction-tuned model (i.e. Qwen2.5-7B-Instruct, Qwen2.5-14B-Instruct);
(2) S1.1: a reasoning-enhanced model obtained by further training the base model on the S1-1.1K dataset~\citep{muennighoff2025s1simpletesttimescaling};
(3) R1-SFT: a baseline model fine-tuned base model on a curated dataset that includes problems sourced from the cold-start SFT dataset along with all responses generated by Deepseek R1;
(4) R1-SFT-RL: a reinforcement learning model that trained (3) with outcome accuracy rewards on the aforementioned 30K RL dataset; 
(5) Cold-Start: a model fine-tuned based model on the constructed cold-start SFT dataset; 
(6) Cold-Start-RL: a reinforcement learning model based on (5), trained using outcome accuracy rewards on the same 30K RL dataset.

\paragraph{Training Details.}
For cold-start fine-tuning, we employ the ms-swift framework~\citep{zhao2024swiftascalablelightweightinfrastructure}, using a learning rate of 1e-5 and a batch size of 8. For difficulty-aware reinforcement learning, we adopt VeRL~\citep{Sheng_2025}, with all RL experiments conducted under a unified setting. Specifically, we follow DAPO~\citep{yu2025dapoopensourcellmreinforcement} to eliminate KL divergence. The policy model is optimized using the AdamW optimizer with a learning rate of 1e-6, a batch size of 256, and a micro-batch size of 32. 
And we set the value of $\alpha$ and $\beta$ both as 0.5, and the value of $\delta$ as 0.625 during training.
During the rollout phase, 16 responses are sampled per prompt, and the maximum generation length is set to 24K tokens. 
All experiments are conducted on NVIDIA H800 GPUs.

\begin{tcolorbox}[title=Baseline Prompt]
\refstepcounter{boxCounter}            
\label{box:baseline} 
Please reason step by step to answer the following Math Problem, and put your final answer in the format of \verb|\boxed|\{answer\}.
\end{tcolorbox}

\begin{tcolorbox}[title=Difficulty-Aware Prompt]
\refstepcounter{boxCounter}
\label{box:difficulty}
Answer the following math problem, judge the difficulty (Easy/Hard) of given problem and start your response with format: [difficulty here], and put your final answer in the format \verb|\boxed|\{answer\}.
\end{tcolorbox}

\paragraph{Evaluation Settings.}
We evaluate our method on four mathematical datasets that span both easy and challenging problems:
AIME2024~\citep{AoPSAIME}, AIME2025~\citep{AoPSAIME}, MATH500~\citep{hendrycks2021measuringmathematicalproblemsolving}, and GSM8K~\citep{cobbe2021trainingverifierssolvemath}. 
The first two datasets consist of more challenging Math Olympiad-style problems, whereas the latter two primarily contain simpler, grade-school level problems, with GSM8K being the easiest among them.
Since AIME2024 and AIME2025 each contain only 30 samples, we report the average performance over 8 independent runs for these two datasets. All evaluations are conducted using consistent inference hyper-parameters set to a temperature of 0.7 and a top-p value of 0.8.
In this paper, we use a baseline prompt to evaluate all baselines that do not incorporate difficulty-aware reasoning. Furthermore, we design a simple difficulty-aware prompt, derived from the baseline prompt, to support our proposed approach, as detailed in Box~\ref{box:baseline} for baseline prompt and Box~\ref{box:difficulty} for difficulty-aware prompt.
To quantify the effectiveness and efficiency of models, we report two metrics: accuracy (Acc.) and the number of tokens generated in the response (Len.).

\begin{table*}[t]
\centering
\caption{Main results of \ourmodel's adaptive mode. We compute two metrics: Acc.(\%) is average accuracy and Len. is the average generated tokens. 
}
\vspace{0.8em}
\resizebox{0.95\linewidth}{!}{%
\begin{tabular}{lcccccccc}
\toprule[1.5pt]
Model & \multicolumn{2}{c}{AIME2024} & \multicolumn{2}{c}{AIME2025} & \multicolumn{2}{c}{MATH500} & \multicolumn{2}{c}{GSM8K} \\
      & Acc.(\%)$\uparrow$ & Len. $\downarrow$ & Acc.(\%) $\uparrow$ & Len. $\downarrow$ & Acc.(\%) $\uparrow$ & Len. $\downarrow$ & Acc.(\%) $\uparrow$ & Len. $\downarrow$ \\
\midrule
\multicolumn{9}{c}{\emph{Qwen2.5-7B-Instruct}} \\
\midrule
Qwen2.5-7B-Instruct& 11.25& 1805.60 &7.08 &1174.06 &73.00 &628.91 &91.58 &272.93 \\
\hdashline
S1.1-7B &16.67 &19022.60 &18.33 &18302.72 &68.80 &5824.47 &\textbf{91.89} &1838.12 \\
R1-SFT-7B &12.08 & 20767.36 & 13.33 & 20184.36 & 62.00 & 8103.41 & 87.34 & 3336.81 \\
Cold-Start-SFT-7B & 11.25 &5553.00 & 7.92 &6237.49 &71.00 &805.78 & 90.22 & 345.44 \\

\hdashline
R1-SFT-RL-7B&21.25&18778.92&17.50&17924.35&66.80&8421.57&88.93&3896.76 \\
Cold-Start-RL-7B& 18.33 & 16911.63 &14.58  & 15941.76 &73.00 & 3797.60 &90.67 & 369.94 \\
\textbf{\ourmodel-7B} & \textbf{21.25}&16889.50 & \textbf{19.17} &15749.08 &\textbf{74.00} &3195.69&90.98 &349.34 \\

\midrule
\multicolumn{9}{c}{\emph{Qwen2.5-14B-Instruct}} \\
\midrule
Qwen2.5-14B-Instruct&11.67 &1043.20 &10.42 &1136.26 &73.60 &568.28 &93.86 &215.42 \\

\hdashline
S1.1-14B &32.92 &17278.03 &\textbf{29.58} &16167.50 &76.00 &4393.17 &\textbf{95.15} &1460.26 \\
R1-SFT-14B &19.58 &20485.99 &20.42 &18447.68 &65.80 &6473.98 &90.67 &2883.76 \\
Cold-Start-SFT-14B &12.92 &4734.60 &12.50 &6431.17 &76.20 &929.48 &93.63 &287.72 \\

\hdashline

R1-SFT-RL-14B&24.17 &18549.26 &23.33 &18294.04 &70.60 &6397.75 &91.05 &2814.69 \\

Cold-Start-RL-14B&27.50 &17722.25 &24.17 &17088.44 &75.60 &4657.38 &94.01 &378.39 \\
\textbf{\ourmodel-14B} \textbf{ }&\textbf{34.58}  &15173.21 &25.83 &14476.93 &\textbf{79.20} &3209.84&94.09 &316.12 \\

\bottomrule[1.5pt]
\end{tabular}
}
\label{tab:main}
\end{table*}

\subsection{Main Results}
The main results are presented in \tablename~\ref{tab:main}. 
We can have the following observations:

\paragraph{\ourmodel Effectively Balances Performance and Reasoning Budget.}
\ourmodel demonstrates competitive overall performance, outperforming most baselines across four datasets while efficiently managing the reasoning budget. Specifically, compared to the RL baseline R1-SFT-RL based on Qwen2.5-7B-Instruct, \ourmodel-7B achieves comparable accuracy on AIME2024 and further improves accuracy by 1.67\% on AIME2025, 7.20\% on MATH500, and 2.05\% on GSM8K. At the same time, it substantially reduces response lengths by 10.06\%, 12.14\%, 62.05\%, and 91.04\% on these datasets, respectively.
Under the 14B setting, \ourmodel-14B increases accuracy by 10.41\%, 2.5\%, 8.6\%, and 3.04\% on AIME2024, AIME2025, MATH500, and GSM8K, while compressing the reasoning budget by 18.20\%, 20.92\%, 49.83\%, and 88.77\% on the corresponding datasets. Similar trends are also observed when comparing \ourmodel to the S1.1-7B and S1.1-14B baselines.
Overall, these results suggest that \ourmodel can adaptively allocate the reasoning budget according to problem difficulty, thereby effectively balancing reasoning efficiency and effectiveness.

\paragraph{Cold-start Fine-tuning Provides an Effective Foundation for Adaptive Budgeting.}
Unlike R1-SFT, which is trained solely on long-form reasoning traces from Deepseek R1, our cold-start fine-tuning strategy (i.e., Cold-Start-SFT) incorporates a combination of both concise and extended reasoning trajectories. This diverse training regimen enables the model to acquire more efficient reasoning strategies, resulting in substantial reductions in response length by 73.26\%, 69.10\%, 90.06\%, and 89.65\% for AIME2024, AIME2025, MATH500, and GSM8K, respectively, on the 7B model, and by 76.88\%, 54.14\%, 85.64\%, and 90.02\% on the 14B model.
Furthermore, when models initialized from the cold-start checkpoint undergo additional reinforcement learning (i.e., Cold-Start-RL), they demonstrate superior budget control compared to those initialized from R1-SFT (i.e.,R1-SFT-RL), yielding additional reductions in response length of 9.94\%, 11.06\%, 54.91\%, and 91.04\% across the same datasets on the 7B model, and 4.46\%, 6.61\%, 27.20\%, and 86.56\% on the 14B model.
These results underscore the critical role of the cold-start fine-tuning phase in establishing a robust foundation for effective adaptive budgeting in downstream tasks.

\paragraph{Our Reward Design Enhance Both Reasoning Effectiveness and Efficiency.}
Building upon the cold-start fine-tuning model, \ourmodel outperforms Cold-Start-RL, which applies reinforcement learning based solely on outcome accuracy by achieving notable accuracy improvements of 2.92\%, 4.59\%, 1.00\%, and 0.31\% on AIME2024, AIME2025, MATH500, and GSM8K, based on  with the 7B model, and improvements of 7.08\%, 1.66\%, 3.6\%, and 0.08\% with the 14B model with much less token consumption. These accuracy gains are attributable to our reward function design, which incorporates both difficulty estimation calibration and difficulty-aware length adjustments. By introducing these additional reward signals, our approach enables the model to more accurately assess problem complexity and allocate computational resources accordingly, thereby achieving a more refined balance between reasoning effectiveness and efficiency.

\begin{table*}[t]
\centering
\caption{
Comparison of adaptive and controlled reasoning modes in \ourmodel. The easy mode reduces reasoning tokens but at the cost of lower performance, making it suitable for fast-response scenarios. Hard mode uses more tokens and yields better results. The adaptive mode automatically estimates problem difficulty to balance effectiveness and efficiency.
}
\vspace{0.6em}
\resizebox{0.99\linewidth}{!}{%
\begin{tabular}{lcccccccc}
\toprule[1.5pt]
Model & \multicolumn{2}{c}{AIME2024} & \multicolumn{2}{c}{AIME2025} & \multicolumn{2}{c}{MATH500} & \multicolumn{2}{c}{GSM8K} \\
      & Acc.(\%)$\uparrow$ & Len. $\downarrow$ & Acc.(\%) $\uparrow$ & Len. $\downarrow$ & Acc.(\%) $\uparrow$ & Len. $\downarrow$ & Acc.(\%) $\uparrow$ & Len. $\downarrow$ \\
\midrule
\ourmodel-7B (Adaptive) & 21.25&16889.50 & 19.17 &15749.08 &\textbf{74.00} &3195.69&90.98 &349.34 \\

\ourmodel-7B (Easy) & 14.58&1652.42  & 10.00 &896.14&70.80 &652.76&90.75 &314.49 \\

\ourmodel-7B (Hard) & \textbf{21.67}& 17562.94 & \textbf{22.08} & 16114.87 & 71.20 &5960.15&\textbf{92.57} &2058.13\\

\midrule
\ourmodel-14B (Adaptive)&34.58  &15173.21 &25.83 &14476.93 &\textbf{79.20} &3209.84&94.09 &316.12 \\

\ourmodel-14B (Easy) &15.83 &1394.78 &11.25 &985.35 &75.40 &578.37 &\textbf{94.31} &284.83 \\

\ourmodel-14B (Hard) &\textbf{36.25} &15004.55 &\textbf{26.25} &14564.21 &74.40 &5321.25 &93.33 & 2598.28\\

\bottomrule[1.5pt]
\end{tabular}
}
\label{tab:control}
\end{table*}

\begin{table*}[t]
\centering
\caption{
Hyper-parameter Analysis.$\alpha$ controls the reward weight for difficulty-estimation calibration, while $\beta$ corresponds to the reward weight for the difficulty-aware length objective. The parameter $\delta$ serves as a predefined threshold for distinguishing between easy and hard samples.
}
\vspace{0.6em}
\resizebox{\linewidth}{!}{%
\begin{tabular}{lcccccccc}
\toprule[1.5pt]
\textbf{\ourmodel-7B} & \multicolumn{2}{c}{AIME2024} & \multicolumn{2}{c}{AIME2025} & \multicolumn{2}{c}{MATH500} & \multicolumn{2}{c}{GSM8K} \\
      & Acc.(\%)$\uparrow$ & Len. $\downarrow$ & Acc.(\%) $\uparrow$ & Len. $\downarrow$ & Acc.(\%) $\uparrow$ & Len. $\downarrow$ & Acc.(\%) $\uparrow$ & Len. $\downarrow$ \\
\midrule
 $\alpha=0.5, \beta=0.5, \delta=0.625$& 21.25&16889.50 & 19.17 &15749.08 &74.00 &3195.69&90.98 &349.34 \\
  $\alpha=1.0, \beta=1.0, \delta=0.625$& 23.75&14762.78 &20.00&13986.64 &73.80&3134.37&90.30&325.04 \\
  $\alpha=0.5, \beta=1.0, \delta=0.625$& 22.08&15616.19&18.75&16116.17&73.80&2836.52&91.13&324.10 \\
  $\alpha=1.0, \beta=0.5, \delta=0.625$& 21.67&17843.66&22.08&16684.28&73.40&3727.77&91.66&365.03 \\
  \hdashline
  $\alpha=0.5, \beta=0.5, \delta=0.5$& 22.92&13790.53&20.42&14040.20&74.00&2888.39&90.60&388.10 \\
  $\alpha=0.5, \beta=0.5, \delta=0.375$& 20.00&13385.01&20.00&13689.74&74.60&2181.96&90.60&323.27 \\

\bottomrule[1.5pt]
\end{tabular}
}
\label{tab:hyper}
\end{table*}

\paragraph{Human-in-the-Loop Controllability of \ourmodel.}
\ourmodel introduces difficulty-aware tags that serve as prerequisite tokens during generation, offering explicit control over response length. To assess the controllability of this mechanism, we emulate user intent by specifying either the “[Easy]” or “[Hard]” tag, which correspond to simplified and complex reasoning modes, respectively. As presented in \tablename~\ref{tab:control}, the experimental results demonstrate that our approach affords effective control over the reasoning budget.
Compared to the adaptive reasoning mode, where the model autonomously determines the problem's difficulty, enforcing the “[Easy]” mode consistently leads to reduced performance across all four datasets. However, it also achieves a substantial decrease in response length by 90.22\% and 94.31\% on the more challenging AIME2024 and AIME2025 datasets for the 7B model, and by 90.81\% and 93.19\% for the 14B model, respectively. Conversely, the “[Hard]” mode enhances performance on most datasets and markedly increases response length, with gains of 86.51\% and 489.15\% on the simpler MATH500 and GSM8K datasets under the 7B setting, and increases of 65.78\% and 721.93\% under the 14B setting.
These findings indicate that \ourmodel enables precise, human-in-the-loop control over reasoning budgets.

\subsection{Analysis}

\begin{table}[t]
\centering
\caption{Ablation study based on 7B model. $r_f$ is the difficulty-estimation calibration reward and $r_l$ is the difficulty-aware length reward.}
\vspace{0.5em}
\resizebox{0.7\linewidth}{!}{%
\begin{tabular}{l|cccc}
\toprule[1.5pt]

Acc.(\%)& {AIME2024}& {AIME2025} & {MATH500} & {GSM8K} \\
\midrule
AdaCtrl-7B & \textbf{21.25} & \textbf{19.17}  &\textbf{74.00} &\textbf{90.98} \\
w/o $r_f$ &17.08 (-4.17\%) & 16.25 (-2.92\%) & 72.20 (-1.80\%) & 89.92 (1.06\%) \\
w/o $r_l$ & 15.42 (-5.83\%) & 16.67 (-2.50\%) & 68.60 (-5.4\%) & 90.60 (-0.38\%) \\

\bottomrule[1.5pt]
\end{tabular}
}
\label{tab:aba}
\vspace{0.7em}
\end{table}

\paragraph{Hyper-parameter Analysis.}
To assess the robustness of our proposed method, we conduct two groups of hyperparameter analyses, as summarized in \tablename~\ref{tab:hyper}. Specifically, we examine: (1) the weights $\alpha$ and $\beta$, which correspond to the difficulty-estimation calibration reward and the difficulty-aware length reward, respectively; and (2) the difficulty threshold $\delta$.
For reward weights analysis, we investigate the effect of varying the reward weights across different combinations $\{1:0.5:0.5, 1:1:1, 1:0.5:1, 1:1:0.5\}$. We observe that accuracy remains largely stable while output length exhibits slight variations. For example, on AIME2024, accuracy stays within 21–23\% as the response length ranges from 14K to 17K tokens. Similarly, on GSM8K, accuracy consistently falls between 90–91\%, while the response length varies from 325–365 tokens.
For difficulty threshold analysis, we further sweep the difficulty threshold $\delta$ over $\{0.625,, 0.5,, 0.375\}$. A smaller value of $\delta$ categorizes more samples as Easy, leading to shorter outputs while maintaining comparable accuracy. For instance, on MATH500, the response length decreases from 3195.69 to 2888.39 and then to 2181.96 tokens, while accuracy remains around 74\%.
These results collectively demonstrate the robustness of our proposed approach under different hyperparameter settings.

\begin{figure}[t]
    \centering
\includegraphics[width=\linewidth]{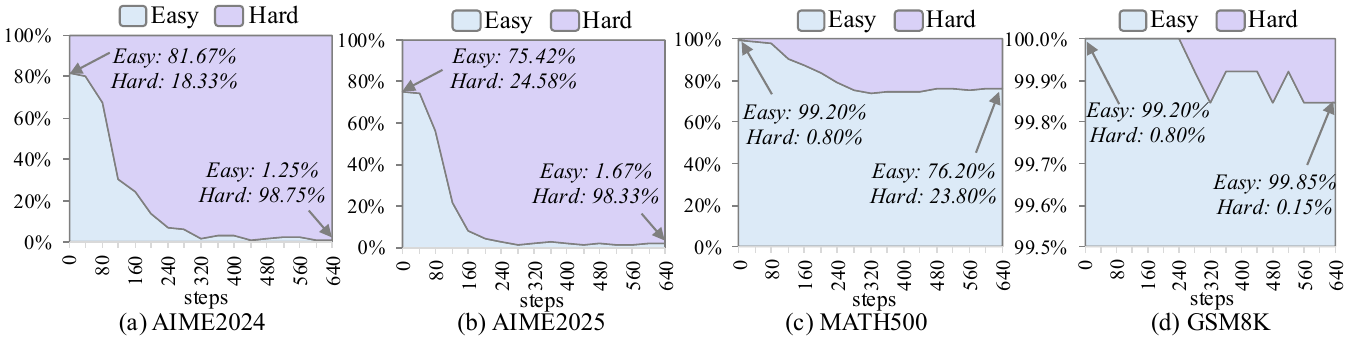}
    \caption{The proportion dynamics of difficulty-aware tags across different datasets during reinforcement learning.}
    \label{fig:prop}
\end{figure}

\paragraph{Ablation Study.}
To further assess the effectiveness of the proposed reward functions, we conducted ablation studies by individually removing the difficulty estimation calibration reward $r_f$ and the difficulty-aware length reward $r_l$. As shown in \tablename~\ref{tab:aba}, the removal of either reward leads to a noticeable decline in performance across all four datasets. These results demonstrate that the combined use of all designed rewards is essential for achieving better optimization.

\begin{figure}[t]
    \centering
\includegraphics[width=\linewidth]{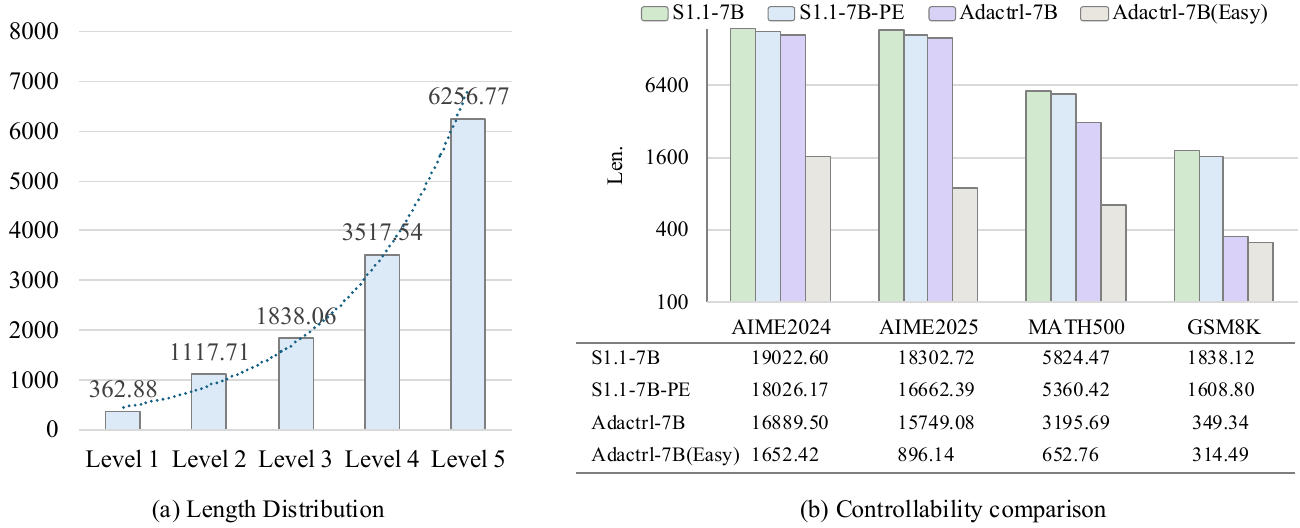}
    \caption{(a) The length of response in different difficulty levels of problems in MATH500, where higher levels indicate more challenging problems; (b) Controllability comparison of AdaCtrl and prompting-based approach.}
    \label{fig:alloacte}
\end{figure}

\paragraph{\ourmodel Serves as Good Difficulty Estimator.}
To better assess the difficulty estimation capability of \ourmodel, we analysis the proportion of difficulty-aware tags generated by \ourmodel-7B across four datasets during  reinforcement learning (RL). As illustrated in \figurename~\ref{fig:prop}, we first observe that, at the initial stage, the model tends to classify most samples in all datasets as easy. This is likely because the cold-start SFT primarily teach the model to explicitly generate self-aware difficulty tags in the expected format, rather than to accurately assess problem difficulty from its own perspective.
However, following our designed difficulty-aware RL, \ourmodel predominantly classifies the majority of problems in the AIME2024 and AIME2025 datasets as hard. These datasets consist of challenging math Olympiad-level problems. In contrast, in the MATH500 dataset, which contains a mixture of easy and difficult problems (the majority of which are relatively solvable by current large language models), the model identifies 76.2\% of problems as easy. For GSM8K, the simplest dataset among the four, over 99\% of problems are categorized as easy, which also accounts for the superior performance of AdaCtrl-14B (Easy) on GSM8K as reported in Table~\ref{tab:control}. These results align with the actual difficulty levels of the datasets and demonstrate that \ourmodel develops a robust capability to estimate problem difficulty through RL.

\paragraph{\ourmodel Facilitates Accurate Difficulty-Aware Budgeting.}
To further investigate the adaptive difficulty-aware budgeting capabilities of \ourmodel, we analyze \ourmodel-7B's responses on the MATH500 dataset, which provides difficulty level annotations for each problem. As illustrated in \figurename~\ref{fig:alloacte} (a), \ourmodel generates progressively longer and more elaborate responses as the difficulty level increases from 1 to 5, ranging from approximately 0.3K to 6K tokens. This trend indicates that \ourmodel can accurately regulate its reasoning budget based on its self-assessed estimation of problem difficulty, thereby enabling automatic and adaptive allocation of computational resources.

\paragraph{Controllability Analysis.}
Previous approaches often rely on vanilla prompting methods to achieve token-level control over reasoning budgets. 
To assess the controllability of \ourmodel, we compare it with a prompting-based baseline that uses the S1.1-7B model as the backbone and follows \citet{nayab2025concisethoughtsimpactoutput} to augment user prompts with the instruction: "Limit the length of the answer to 500 tokens." 
The response lengths generated by \ourmodel in both easy and adaptive modes, as well as those produced by S1.1-7B and its prompting-based budget control variant, are reported in \figurename~\ref{fig:alloacte} (b).
We can observe that, through the prompting-based method (i.e., S1.1-7B-PE) explicitly restricting outputs to 500 tokens, it only yields reductions in response length of only 5.25\%, 8.96\%, 7.97\%, and 12.48\% on the AIME2024, AIME2025, MATH500, and GSM8K datasets, respectively, falling significantly short of the targeted 500-token limit. We also observed similar trends when the length constraint was set to 1000 tokens. These results suggest that achieving precise, fine-grained control over output length is challenging due to highly dependent on model’s instruction-following capabilities, and that current models may struggle to reliably interpret and execute such granular prompts.
In contrast, the easy mode of \ourmodel achieves substantially greater compression of reasoning budgets, reducing response lengths by 90.22\%, 94.32\%, 79.57\%, and 9.98\% across the same datasets. This demonstrates superior controllability, which can be primarily attributed to \ourmodel's mixed fine-tuning strategy and its design of a difficulty-aware length reward.

\begin{wrapfigure}{r}{0.5\textwidth}  
    \centering
    \includegraphics[width=0.49\textwidth]{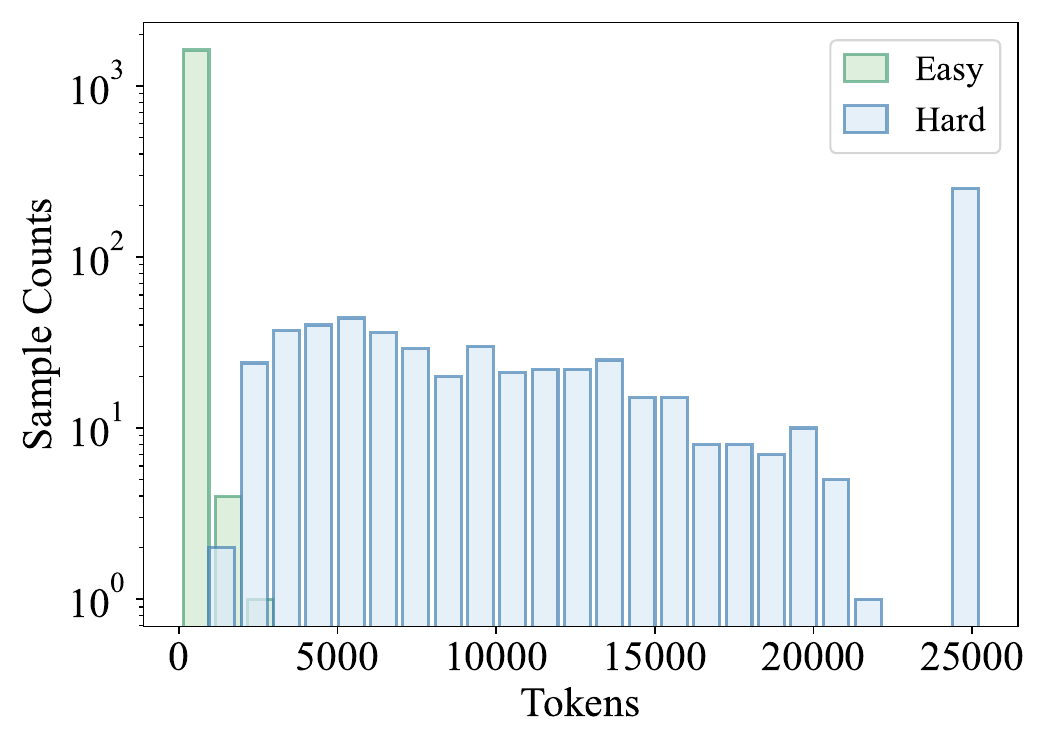}
    \caption{The length distribution of response generated by \ourmodel-7B.}
    \label{fig:ada_len_7b}
\end{wrapfigure}

\paragraph{Length Distribution of Generated Responses.}
To further demonstrate the budget allocation capability of \ourmodel, we analyze the response length distribution of \ourmodel-7B across all inference results from the evaluated benchmarks.
Specifically, we compute the distributions separately for samples categorized as easy and hard by the model predicted tags.
As shown in \figurename~\ref{fig:ada_len_7b}, \ourmodel clearly differentiates between easy and hard problems. Notably, the response lengths for easy problems are concentrated within a relatively narrow range.
These findings indicate that our approach achieves effective and accurate budget control guided by the self-assessed difficulty-aware tags.

\begin{figure}[t]
    \centering
\includegraphics[width=\linewidth]{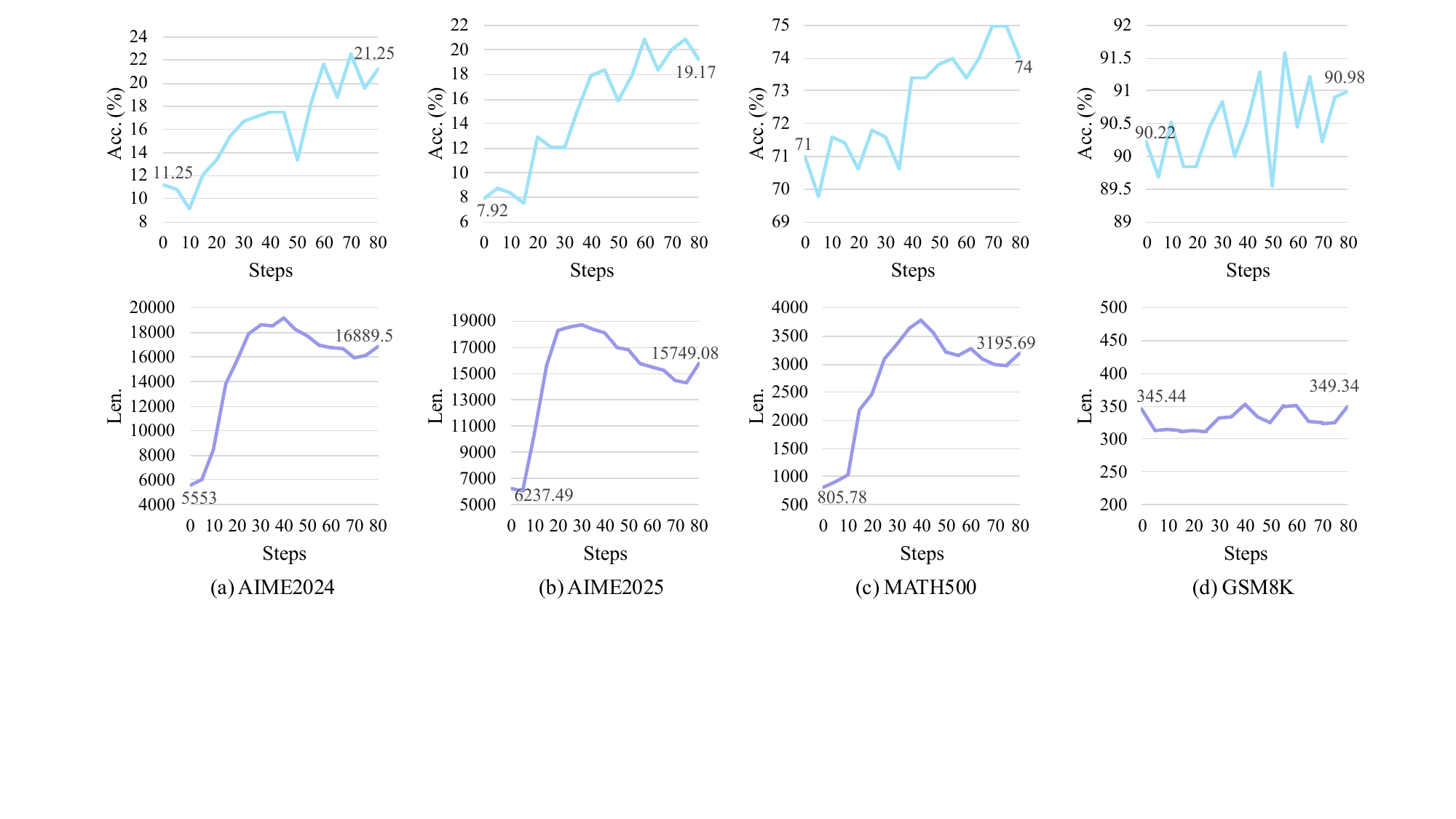}
    \caption{Training dynamics of accuracy and response length across different datasets during reinforcement learning.
    }
    \label{fig:rl}
\end{figure}

\paragraph{Training Dynamics During RL.}
To further investigate how the model learns to allocate adaptive reasoning budgets, we analyze both performance trends and budget dynamics of \ourmodel-7B throughout the reinforcement learning (RL) training process. As illustrated in Figure~\ref{fig:rl}, the model exhibits upward trends in performance across all four datasets, suggesting a progressive enhancement in reasoning capabilities.
Regarding budget dynamics, we observe distinct patterns across datasets. On AIME2024, AIME2025, and MATH500, the average response length initially increases rapidly during early training steps, then gradually decreases and stabilizes at a level longer than that before RL training.
While for GSM8K, the response length remains relatively stable and close to that observed before RL training. 

These findings suggest that the reasoning budget allocation learned during cold-start fine-tuning is insufficient for more complex problems, such as those in AIME2024, AIME2025, and MATH500. Consequently, the model adjusts its budget dynamically in response to actual problem difficulty during RL phase. In contrast, for the comparatively simpler GSM8K dataset, the model is already capable of effectively allocating minimal budgets after cold-start fine-tuning, indicating its ability to distinguish and handle easier problems without requiring significant adjustment.

\section{Conclusion}
In this work, we propose an adaptive and controllable reasoning framework designed to mitigate the problem of overthinking while granting users explicit control over computational resources. To this end, we introduce \ourmodel that supports both dynamic reasoning budget allocation and user-directed budget adjustments. Our approach utilizes a two-stage training pipeline that combines cold-start fine-tuning with difficulty-aware reinforcement learning. Experiments conducted on four benchmark datasets demonstrate that \ourmodel effectively allocates reasoning budgets based on self-assessed problem difficulty, leading to performance improvements while dynamically reducing response lengths by 10\%–90\%. This enables flexible trade-offs between efficiency and performance. Furthermore, \ourmodel unlocks the potential of human-in-the-loop control towards reasoning budgets according to tailored needs.

\bibliographystyle{plainnat}
\bibliography{ref}

\end{document}